\documentclass{article}
\usepackage{amsmath,amssymb,amsfonts,spconf}
\usepackage{enumitem}
\setlist{nosep, leftmargin=14pt}

\usepackage{cite}
\usepackage{pifont}
\usepackage{algorithmic}
\usepackage{graphicx}
\usepackage{textcomp}
\usepackage{xcolor}
\usepackage{xspace}
\usepackage[colorlinks=true,linkcolor=blue,citecolor=blue, urlcolor=blue]{hyperref}
\def\BibTeX{{\rm B\kern-.05em{\sc i\kern-.025em b}\kern-.08em
    T\kern-.1667em\lower.7ex\hbox{E}\kern-.125emX}}
\usepackage{tabularray}

\usepackage[normalem]{ulem} 

\definecolor{commentred}{RGB}{240,0,0} 

\def\sd#1{\color{black} {\scriptsize $\pm$ \hspace{-1.4mm} #1} \color{black}}

\newcommand{\xmark}{\ding{53}}%
\newcommand{\xmarkk}{\ding{55}}%
\newcommand{\mname}{SHMC-Net\xspace}

\begin{document}

\title{SHMC-NET: A Mask-guided Feature Fusion Network for 
Sperm Head Morphology Classification}



\name{Nishchal Sapkota$^{\dagger}$, \quad Yejia Zhang$^{\dagger}$, \quad Sirui Li$^{\dagger}$,\quad Peixian Liang$^{\dagger}$,\quad Zhuo Zhao$^{\dagger}$,\vspace{-0.48cm}}
\address{\textit{Jingjing Zhang}$^{\star}$,\quad \textit{Xiaomin Zha}$^{\star}$,\quad \textit{Yiru Zhou}$^{\star}$,\quad \textit{Yunxia Cao}$^{\star}$,\quad \textit{Danny Z. Chen}$^{\dagger}$ \vspace{2mm}\\$^{\dagger}$Department of Computer Science and Engineering, University of Notre Dame, IN 46556, USA \\ $^{\star}$First Affiliated Hospital of Anhui Medical University, Obstetrics and Gynecology, Hefei, Anhui, China \vspace{-0.45cm}}
\maketitle
\begin{abstract}

Male infertility accounts for about one-third of global infertility cases. Manual assessment of sperm abnormalities through head morphology analysis
encounters issues of observer variability and diagnostic discrepancies among experts. Its alternative, Computer-Assisted Semen Analysis (CASA), suffers from low-quality sperm images, small datasets, and noisy class labels. 
We propose a new approach for sperm head morphology classification, called \textit{\mname},
which uses segmentation masks of sperm heads 
to guide 
morphology classification of sperm images.
\mname generates reliable 
segmentation masks using image priors, refines object boundaries with an efficient graph-based method,
and trains an image network with sperm head crops and a mask network with the corresponding masks. In the intermediate stages of the networks, image and mask features are fused with a fusion scheme to better learn morphological features. 
To handle noisy class labels and regularize training on small datasets, \mname applies \textit{Soft Mixup} to combine 
mixup 
augmentation and a loss function.
We achieve state-of-the-art results on SCIAN and HuSHeM datasets. Code is available in \href{https://github.com/nsapkota417/SHMC-Net}{GitHub}. 
\end{abstract}


\section{Introduction}


Approximately 1 in 6 adults worldwide experience infertility \cite{world2023infertility}, and nearly one-third of these cases are attributed to male infertility.
Human sperm head morphology classification is clinically significant because different types of sperm head morphology abnormalities are indicative of genetic or environmental factors that impact treatment decisions~\cite{menkveld2011measurement}.
However, manual classification of these abnormalities is 
subjective, 
time-consuming, 
and heavily dependent on human experience, which leads to observer variability \cite{morphSubjectivity} and diagnostic discrepancies (noisy class labels) even among experts.
Away from subjective manual assessments, computational approaches such as Computer Assisted Semen Analysis (CASA) have advanced morphology analysis toward a more objective and reproducible process~\cite{whoHandbook}.
Traditional CASA methods for sperm morphology classification
used hand-crafted shape descriptors and yielded competitive results compared to manual classification \cite{vchang2014goldSeg, violetaCESVM, shakerAPDL,goldScian}. 
Recent advances in Computer Vision (CV) and Deep Learning (DL) have developed methods~\cite{liuTL,charleyCasa} that no longer require hand-crafted features and generally outperform traditional methods.



\begin{figure}[t] 
\begin{center} 
\includegraphics[width=0.8\linewidth]{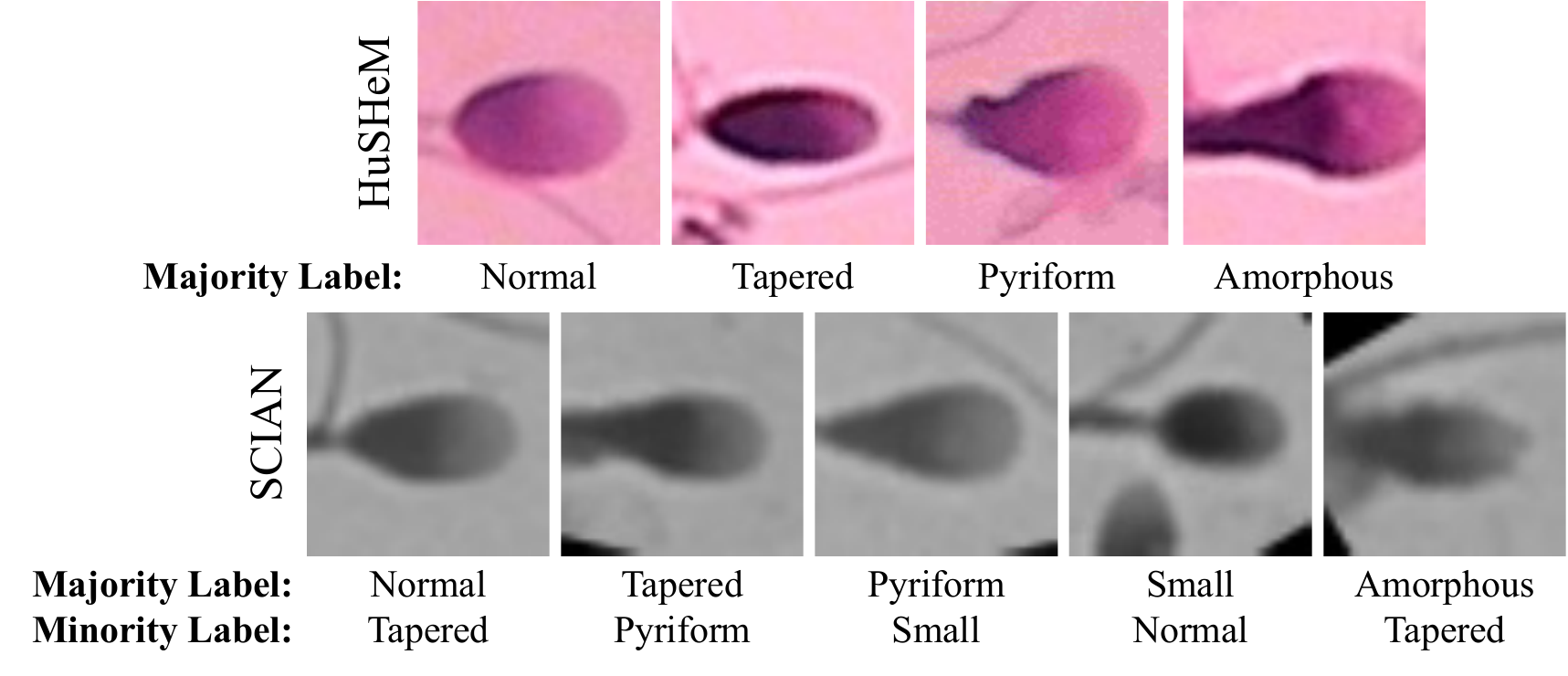} 
\vspace{-3mm}
\caption{Row 1: Sperm head crops with different labels (HuSHeM). Row 2:  Class label discrepancies (SCIAN).\vspace{-10mm}}
\label{example-samples} 
\end{center}
\end{figure}

Existing sperm datasets often contain a limited quantity of images, which may cause overfitting or over-focusing of DL models on clinically irrelevant image regions \cite{charleyCasa}. 
The labels of such images may be noisy or inconsistent, due to the subjective nature of expert annotations. 
(e.g., see row 2 of Fig.~\ref{example-samples}).
Sperm images 
are often of low quality and have various
artifacts in the background, varying textures in the foreground, and irregular structures 
(e.g., in the $2^{nd}$ or $5^{th}$ crop of row 2 of Fig.~\ref{example-samples}, 
the thick sperm mid-section could be taken as part of the head), 
which can affect DL models to learn and discriminate effectively, especially with a limited number of samples and noisy labels. 

\begin{figure*}[h!] 
\begin{center} 
\setlength{\abovecaptionskip}{-2cm}   
\setlength{\belowcaptionskip}{-2cm}   
\includegraphics[width=0.7\textwidth]{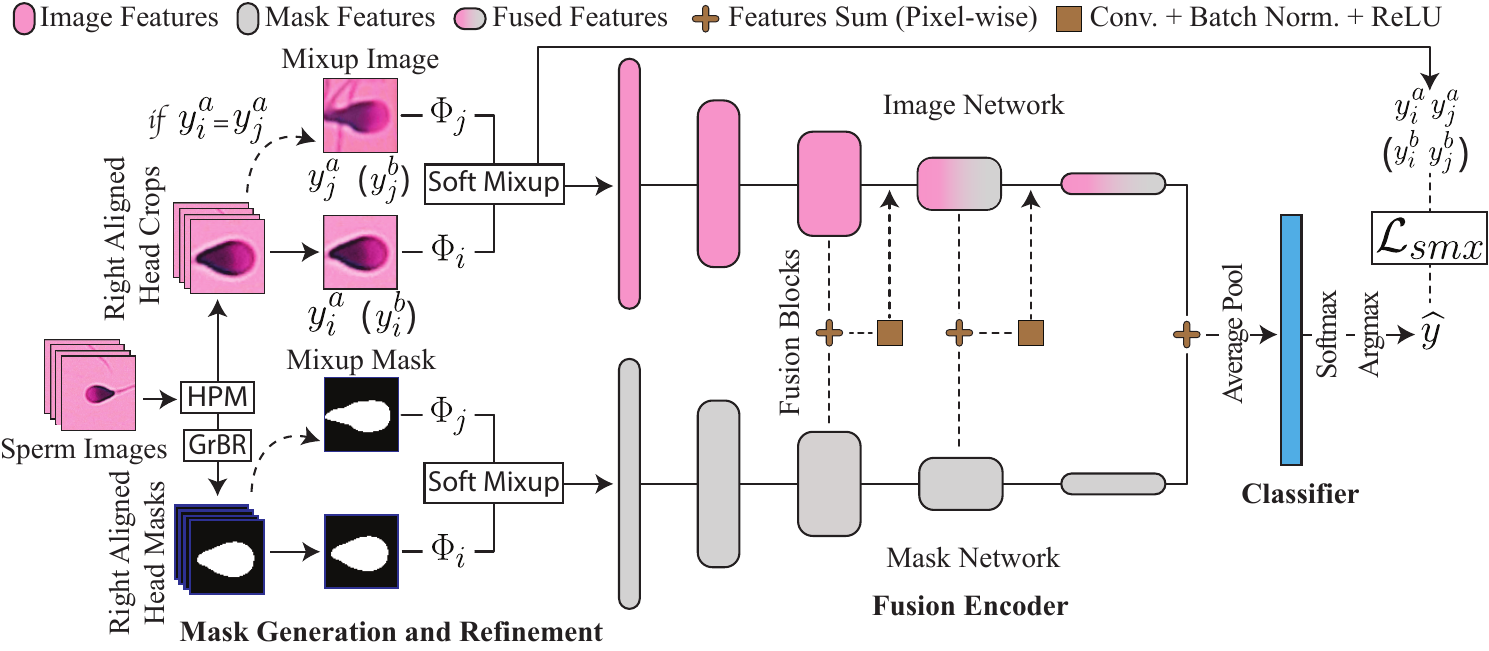} 
\caption{\small{An overview of our \mname.
}}
\vspace{-0.7cm}
\label{main diagram} 
\end{center}
\end{figure*} 
To address these challenges we 
propose to incorporate additional information from sperm head masks 
to help morphology classification.
Sperm head masks can be reliably and efficiently extracted from images, as shown in previous work \cite{park1997segmentation,vchang2014goldSeg,charleyCasa, ilhan2019automaticEllep}. 
Such masks in general have well-delineated boundaries with fewer background artifacts. 
The objects of interest in the masks are represented as the foreground with morphologically relevant shape, size, and structure information that is easy to extract while containing little distracted structures that may inhibit shape analysis.  
Motivated by these observations, we propose a new approach for effective {\bf S}perm {\bf H}ead {\bf M}orphology {\bf C}lassification, called {\bf \mname}. 
Our major contributions are 
(1) a new sperm head morphology-aware classification network architecture, \textit{Fusion Encoder}, that applies an effective feature fusion scheme to utilize morphology information from both the input images and segmentation masks  (3) a novel combination of augmentation and loss function to handle noisy labels and regularize training on small datasets, (4) an efficient graph-based sperm head mask boundary refinement method and (5) SOTA performances on 2 public sperm datasets 
outperforming even the best-known methods that use additional pre-training or costly ensembling of multiple convolution neural network (CNN) models.



\section{Methodology}
\vspace{-0.2cm}

Fig.~\ref{main diagram} gives an overview of our proposed \textbf{S}perm \textbf{H}ead \textbf{M}orphology \textbf{C}lassification \textbf{N}etwork, \mname.
Given a raw sperm image, we first apply our \textit{mask generation and refinement} module to obtain sperm head crops and their corresponding boundary-refined masks. 
The head crops and their corresponding masks are fed into {\it image network} and {\it mask network} of the \textit{Fusion Encoder}, respectively. 
In the deeper stages of the two networks, image and mask features are fused
and passed into the subsequent stages of the image network. 
The final features obtained from both networks are fused and passed through a linear classifier to obtain the class predictions. 
\mname also uses \textit{Soft Mixup}, which consists of intra-class Mixup augmentation for both images and masks and a loss function, to address the observer variability and class imbalance problems and regularize the training on small datasets. 
The three major components of \mname  --- mask generation and refinement, Fusion Encoder, and Soft Mixup regularization --- are detailed below.

\vspace{-0.2cm}
\subsection{Mask Generation and Refinement} \label{Mask Generation}
\vspace{-0.1cm}
Reasonable head masks can be extracted from sperm images
in an unsupervised fashion \cite{park1997segmentation,vchang2014goldSeg,charleyCasa, ilhan2019automaticEllep}
but these masks may have imperfect boundaries. 
Hence, given a raw sperm image, we first generate segmentation masks for sperm head crops using anatomical and image priors and further refine mask boundaries.
Using the HPM method \cite{charleyCasa}, we obtain the sperm-head-only crops from the sperm image, right-align them, and produce their pseudo-masks. 
We propose a sperm head shape-aware Graph-based Boundary Refinement (GrBR) method. 
GrBR generates the boundary-refined masks instantaneously ($<7$ ms per image).
The GrBR module refines the boundary contour $C$ of the HPM pseudo-mask in a head crop to obtain a closed curve $C'$ that captures the accurate head boundary of the mask. We formulate the optimal contour refinement problem as computing the shortest path in a directed graph $G$ defined on $C$. We first uniformly sample $n$ points clockwise on $C$, and for each sample point $a_j\in C$, a line segment $l(a_j)$ is drawn orthogonal to the curvature of $C$ 
at $a_j$, with $a_j$ being the center point of $l(a_j)$. We then sample $m$ points uniformly on each line segment $l(a_j)$. Let $p_i^j :=(x(p_{i}^{j}), y(p_{i}^{j}))$ represent the $i$-th sample point on the line segment $l(a_j)$, $i=1,2,\ldots,m$. All the sample points on $C$ and on the line segments are vertices of the graph $G$ and have weights equal to the negatives of their gradients. Since sperm heads have almost convex shapes, we impose a smoothness constraint and a near convex constraint for computing $C'$.
The smoothness constraint is enforced by using a smoothness parameter $s$ such that each $p_i^j$ can have a directed edge to any point $p_{i'}^{j+1-\lfloor j/n \rfloor \times n}$ only,
for each $i'$ with $|i'-i|\leq s$, where  $i=1, 2, \dots, m$, $i'=1, 2, \dots, m$, and $j=1, 2, \dots, n$. 
The shape constraint is enforced by penalizing any concave edge-to-edge connection along $C'$ with a larger connection cost, i.e., edge weight = $c\times(\theta-\pi)$ if $\theta > \pi$, and else edge weight = 0, where $c$ = cost penalty = initial edge weight
and $\theta$ is the angle between the two adjacent edges. We make $G$ into a directed acyclic graph by making a copy of all the vertices on the first line segment $l(a_1)$. 
These vertex copies are now on the line segment copy $l'(a_1)$ and their edges are from the vertices of $l(a_n)$.
An optimal closed curve $C'$ in the head crop subject to the smoothness and shape constraints based on $C$ 
is obtained by computing the shortest path in the graph $G$ in polynomial time using dynamic programming. 
\begin{table*}[h]
\begin{center}
\renewcommand\arraystretch{0.1}
\caption{\label{tab: Main Results} \small{Results on HuSHeM datasets.
In the column of Pre-training, IN denotes ImageNet pre-trained weights, US denotes unsupervised pre-training, and SMIDS denotes pre-training on the SMIDS dataset. `--' marks that corresponding results are not available.}}
\vspace{0.1cm}
\resizebox{0.7\textwidth}{!}{
\begin{tblr}{columns={colsep=3pt},
    colspec={r | c  | l l l l | l l l l},
    row{1-2} = {gray!20!black!15},
    row{15} = {gray!20!black!5},
    }
\hline
\multirow{2}{*}{\hspace*{5.4mm} Method} & \multirow{2}{*}{Pre-training} & \multicolumn{4}{c} {SCIAN Partial Agreement (PA)} & \multicolumn{4}{c} {HuSHeM}\\
\cline{3-10} 
& & Accuracy & Precision & Recall & F1 & Accuracy & Precision & Recall & F1 \\
\hline
\hline
CE-SVM \cite{violetaCESVM}& \xmark          & --        & --        & 57.6         & -- &   78.7       & 80.6     & 78.6     & 79.6 \\

ADPL \cite{shakerAPDL}& \xmark   & --        & --        & 62.4         & -- &   92.6       & 92.7     & 92.7     & 92.7 \\

MC-HSH \cite{iqbalMCHSH}& \xmark  & 63.0        & 56.0        &  68.0       & 61.0 & 95.7          &96.1     &95.5     &95.5  \\
Y{\"u}zkat et.al \cite{yuzkat2021multi}& \xmark   &71.9        &\textbf{67.0}        &52.8     &59.1  
                              & 85.2      & 85.2    & 85.3    & 85.3\\
FT-VGG \cite{riordonFTVGG}& IN   & 61.7        & 61.9        &  61.7       & 61.8 & 94.1          &94.3     &94.1     &94.2  \\
TL \cite{liuTL} & IN   &--         &   --      & 62.0         & -- & 96.4          &96.4     &96.4     &96.4  \\

Ilhan et.al\cite{fusionSCIDS}& IN+SMIDS  & 73.2        & 66.2        & 57.6    & 61.6 
                              & 92.1       & 92.3     & 92.1     & 92.2 \\
Zhang et.al\cite{charleyCasa}& IN+US  & 65.9        &58.7           &\textbf{68.9}   & 63.3 
                              & 96.5          & 96.8         &96.6    & 96.6 \\
\hline
\hline
\mname (ours) & IN &\textbf{73.6} \sd{0.4}	&66.5 \sd{0.8}	&63.6\sd{0.9}	&\textbf{65.0} \sd{0.3}
&\textbf{98.2} \sd{0.3}	&\textbf{98.3} \sd{0.3}	&\textbf{98.1} \sd{0.3}	&\textbf{98.2} \sd{0.3}	\\
\hline
\end{tblr}
}
\vspace{-0.75cm}
\end{center}
\end{table*}
\vspace{-0.40cm}
\subsection{Fusion Encoder}
\vspace{-0.1cm}
Using the masks directly to train a classification network is not ideal (see Table \ref{ablation}) because the masks lack texture details and are polarized to all-or-nothing errors due to their coarse discretized representation.
Hence, we propose a novel \textit{Fusion Encoder} that uses a one-directional feature fusion scheme (from mask network to image), at the deeper layers (evidence for this design in Section \ref{fusion_scheme}) to exploit the learned semantics from both networks. 
The Fusion Encoder 
consists of an image network, a mask network, fusion blocks, and a classifier.

\textbf{Image and Mask Network}:
We use two 5 staged lightweight ShuffleNet \cite{shufflenet} backboned encoders as our image and mask network. 
In the first three stages, both networks learn the features independent from each other. After the third stage, features from the image and mask networks are fused and passed to the fourth stage of the image network. The features from the two networks are fused again before the fifth stage and passed to the image network to obtain the final features from the image network. 

\textbf{Fusion Blocks and Classifier} \label{fusionBlock}
Fusion blocks are placed at the beginning of the $4^{th}$ \& $5^{th}$ stages of the networks and consist of a pixel-wise summation of the image and mask features, followed by a convolution, batch normalization, and ReLU operations. 
A pixel-wise summation is performed between the final features obtained at the end of the image and mask networks, followed by average pooling and fed through a randomly initialized fully connected linear classifier that outputs (after softmax \& argmax) the morphology class predictions of the sperm head crop.
\vspace{-0.2cm}
\subsection{Soft Mixup Regularization} \label{softMixup}
\vspace{-0.1cm}
Accurately classifying the sperm head morphology is difficult even for experts, which may lead to disagreement on expert-labeled class IDs of samples (i.e., for SCIAN a sample may have a majority class label with consensus among at least 2/3 expert labels and a minority class label with no consensus).

\textbf{Soft Mixup Augmentation}: 
We apply an intra-class Mixup augmentation \cite{zhang2018mixup} scheme to handle samples with disagreed labels and reduce class conditional input variance. Given a sample image crop $x_i$ and its corresponding mask $m_i$ with a majority label $y^{a}_i$ and a minority label $y^b_i$, we randomly pick another sample $x_j$ (with a mask $m_j$, a majority label $y^a_j$, and a minority label $y^b_j$) if $y^a_i = y^a_j$. Prior to this step, we perform random oversampling of all the under-represented classes 
to match the number of samples in the most represented class. 
We then apply separate image augmentation ($\Phi$) to $x_i$ and $x_j$.
The input, $\ddot{x}_i$, to the Fusion Encoder is a linear combination of the augmented mixup samples $\Phi_i(x_i)$ and $\Phi_j(x_j)$, and its target label $\ddot{y}_i$ is equivalent to the majority label ($y_i^a$).
Formally:
\begin{equation}
\label{mx-sample}
\begin{aligned}
\ddot{x}_i &:= \lambda \; \Phi_i(x_i) + (1-\lambda) \; \Phi_j(x_j), 
\; \; \ddot{y}_i:= y^a_i = y^a_j, \\
\ddot{m}_i &:= \lambda \; \Phi_i(m_i) + (1-\lambda) \; \Phi_j(m_j).
\end{aligned}
\end{equation}
where $\lambda$ is the mixup strength. 
Soft Mixup augmentation yields a mixed crop $\ddot{x}_i$, mixed mask $\ddot{m}_i$, target label $\ddot{y}_i$, and using these as input, the model outputs class prediction $\widehat{y}_i$.



\textbf{Soft Mixup Loss}:
We modify the Soft Loss in~\cite{charleyCasa} 
to adapt intra-class mixup to handle partial agreements among the morphology class labels. 
Soft Loss is defined as:
\vspace{-0.1cm}
\begin{equation} 
\label{mx-Soft-Loss}
\mathcal{L}_{s} (\widehat{y}, y^a, y^b) = \gamma  \mathcal{L}_{ce}(\widehat{y},y^a) + (1-\gamma) \mathcal{L}_{ce}(\widehat{y},y^b),
\vspace{-0.1cm}
\end{equation}
where $\mathcal{L}_{ce}$ is the cross entropy loss and $\gamma$ is a hyper-parameter that weighs the majority label. Then, we combine our soft loss with the intra-class mixup to build the \textit{Soft Mixup Loss},
as:
\vspace{-0.1cm}
\begin{equation} \label{smx}
\mathcal{L}_{smx} = \lambda \mathcal{L}_{s} (\widehat{y}_i, y^a_i, y^b_i) + (1-\lambda)\mathcal{L}_{s} (\widehat{y}_i, y^a_j, y^b_j),
\vspace{-0.1cm}
\end{equation} 
where $\lambda$ is the same as used in Eq.~(\ref{mx-sample}). When minority labels are not available we retain our formulation of $\mathcal{L}_{smx}$ but with $y_i^a = y_i^b$, reducing our loss to that in the original Mixup~\cite{zhang2018mixup}.



   
\vspace{-1mm}
\section{Dataset, Experiments, and Results}
\vspace{-0.05cm}
\subsection{Datasets}
\vspace{-0.1cm}
We demonstrate the performance of \mname on two public sperm morphology datasets.
SCIAN-Morpho Sperm GS (SCIAN) \cite{goldScian} contains 1854 gray-scale sperm images of size  35 $\times$ 35 each belonging to one of five morphological classes: Normal (N), Tapered (T), Pyriform (P), Small (S), or Amorphous (A) as labeled by 3 experts. Samples where all 3 experts agree on the same class label form the SCIAN total Agreement (SCIAN-TA) dataset (384 images: 35 N, 69 T, 7 P, 11 S, 262 A), and the samples where at least 2 experts agree on class labels form the SCIAN Partial Agreement (SCIAN-PA) dataset (1132 images: 100 N, 228 T, 76 P, 72 S, 656 A)

Similarly, 
the Human Sperm Head Morphology (HuSHeM) dataset \cite{shakerAPDL} contains 216 RGB sperm images of size 131 $\times$ 131 
belonging to one of the 
4 morphological classes: Normal (54), Tapered (53), Pyriform (57), and Amorphous (52).



\vspace{-0.2cm}
\subsection{Implementation Details} 
\vspace{-0.1cm}
ImageNet pre-trained ShuffleNet V2 \cite{shufflenet} is used as the backbone for our image and mask networks and 
trained using the AdamW optimizer 
with an initial learning rate of 0.00015 and the Cosine Annealing learning rate scheduler. 
The model is trained 
on a single NVIDIA TITAN-Xp GPU 
for 300 epochs (batch size of 128 for SCIAN-PA and 64 for SCIAN-TA and HuSHeM). 
Dropout with 40\% probability and weight decay of 0.01 was applied as additional regularization. 
The majority class weight $\gamma$ is set to 0.85 and mixup strength, $\lambda$, is sampled randomly from the beta distribution with $\alpha = 0.5$. All the image crops are first resized to 64$\times$64 and 
randomly square-cropped between sizes 45 to 64 and resized again to 64*64 along with 10-degree rotation, vertical flip, and horizontal/vertical shift by up to 10\% of their width or height. 
We evaluate the performance in terms of Overall \textit{Accuracy}, macro-averaged class-wise \textit{Precision} and \textit{Recall}, and the \textit{F1} score attained as a harmonic mean of the two. We apply 5-fold cross-validation and report the average of 5 runs.


\vspace{-0.2cm}
\subsection{Performance Comparison on SCIAN}
\vspace{-0.1cm}
As seen in Table~\ref{tab: Main Results},
our \mname yields the highest Accuracy and F1 on SCIAN-PA. 
Known SOTA methods traded Precision and Recall, 
while \mname balances these two and gives the best F1. 
Similarly, for SCIAN-TA, we achieve the best Accuracy, Precision, and F1 
outperforming even the ensemble methods (Table \ref{additionalResults}).

\vspace{-0.2cm}
\subsection{Performance Comparison  on HuSHeM}
\vspace{-0.1cm}
On HuSHeM dataset \mname outperforms the previous SOTA methods on all the metrics despite the diminished margins for performance improvement (Table~\ref{tab: Main Results}). 
We also implement an ensemble version of \mname (\mname-E) using three backbones (ShuffleNet V2, ResNet34 \cite{resnet}, and DenseNet121 \cite{densenet}), and apply 
majority voting.
With 6 times fewer parameters ($\sim$50 millions) SHMC-Net-E  outperforms \cite{spencer2022ensembled} ($\sim$300 millions) by about 0.65\% on all metrices.


\vspace{-0.2cm}

\subsection{Component Studies} \label{fusion_scheme}
We study different fusion directions: image network to mask (I2M), mask to image (M2I), and bidirectional (BD) where both networks mutually help each other and no fusion (\xmarkk). 
At the classifier level, we explore the options to use features from either the image network (I) or both networks (B). In Table~\ref{Components Study}, we observe that 
avoiding fusion in the early stages, using mask features for fusion (M2I) in the later stages, and fusing both the image and mask features before the classifier offers the best performance.

\begin{table}
\begin{center}
\caption{\label{additionalResults} \small{Additional Results on SCIAN-TA and HuSHeM.}}
\vspace{0.2cm}
\resizebox{0.45\textwidth}{!}{
\begin{tblr}{
    colspec={r | X[1.6cm,l] X[1.6cm,l] X[1.6cm,l] X[1.6cm,l]},
    row{1,2,8} = {gray!20!black!15},
    row{7,10} = {gray!20!black!5}
}
\hline
Method  & Accuracy & Precision & Recall & F1 \\
\hline  
\hline
\multicolumn{5}{l}{SCIAN Total Agreement (TA)} \\
\hline
MC-HSH \cite{iqbalMCHSH}    & 77.0   & 64.0  & 88.0   & 74.0 \\
Spencer et al.~\cite{spencer2022ensembled}   & 86.3 & 77.0  & \textbf{91.8}  & 81.2 \\
\hline
\mname (ours)  &\textbf{86.9} \sd{0.8} 
&\textbf{86.7} \sd{1.4}
&83.4 \sd{1.6}
&\textbf{85.0} \sd{1.5}\\
\hline
\hline
\multicolumn{5}{l}{Ensemble for HuSHeM} \\
\hline
Spencer et al.~\cite{spencer2022ensembled}  & 98.52 & 98.53  & 98.52  & 98.52 \\
\hline
\mname-E (ours)   & \textbf{99.17} \sd{0.20} & \textbf{99.19} \sd{0.21} & \textbf{99.16} \sd{0.20} & \textbf{99.17} \sd{0.20}\\
\hline
\end{tblr}
}
\caption{\label{Components Study} \small{Study of key components.}}
\resizebox{0.35\textwidth}{!}{
\begin{tblr}{columns={colsep=3pt},
    colspec={l| c c| c c },
    row{1-2,8,11} = {gray!20!black!15},
    row{12} = {gray!20!black!5}
}
\hline
\multirow{2}{*}{Fusion Options} & \multicolumn{2}{c} {SCIAN-TA} & \multicolumn{2}{c} {HuSHeM}\\
\cline{2-10} 
& Accuracy & F1 & Accuracy & F1 \\
\hline
\hline
M2I M2I M2I M2I B &62.9&58.5&90.7&91.2 \\
BD BD BD BD B  &64.0&58.4&90.7&91.1 \\ 
I2M I2M M2I M2I B &70.4  &62.9 &97.2 &97.3 \\
\xmarkk \; \xmarkk \; M2I M2I I &72.3&64.5&97.7&97.8 \\
\hline
\end{tblr}
}
\caption{\label{ablation}{\small Ablation study results. C.L. = classifier level, I.S. = intermediate stage fusion.}}
\vspace{0.2cm}
\resizebox{0.49\textwidth}{!}{
\begin{tblr}{
    colspec={l l l l l | c c | c c },
    row{1-2} = {gray!20!black!15},
}
\hline
\multirow{2}{*}{\rotatebox{20}{Image}} & 
\multirow{2}{*}{\rotatebox{20}{Masks}} & 
\multirow{2}{*}{\rotatebox{20}{C.L. Fusion}} & 
\multirow{2}{*}{\rotatebox{20}{I.S. Fusion}} & 
\multirow{2}{*}{\rotatebox{20}{Soft Mixup}} & 
\multicolumn{2}{c} {SCIAN-PA} & \multicolumn{2}{c} {HuSHeM}\\
\cline{5-14} 
& & & & &  Accuracy & F1 & Accuracy & F1 \\
\hline  
 \checkmark &&&&               &70.0 &57.5 &94.5         &94.7  \\
    & \checkmark &&&    &58.8 &43.9 &82.0        & 82.2  \\
    \hline
  \checkmark &\checkmark& \checkmark&&  &71.6& 60.9& 96.3  & 96.5   \\
  \checkmark &\checkmark& \checkmark& \checkmark&  &72.5 & 62.8 &  97.7  & 97.8  \\  
\checkmark &\checkmark& \checkmark& \checkmark&\checkmark      &\textbf{73.6}& \textbf{65.0}& \textbf{98.2}  & \textbf{98.2}  \\ 
\hline
\end{tblr}
}
\vspace{-0.6cm}
\end{center}
\end{table}


\vspace{-0.4cm}
\subsection{Ablation Study}
\vspace{-0.1cm}
Combining mask and image features at the classifier level has improvements (3.4\% F1 on SCIAN-PA and 1.8\% on HuSHeM) over using image-only. Intermediate stage fusion of mask and image features further strengthens the model's ability to learn morphology features (further 1.9\% on SCIAN-PA and 1.3\% on HuSHeM). 
Finally, handling class imbalance and label discrepancies through Soft Mixup adds 2.2\% further F1 improvement on SCIAN-PA, and the regularization through Soft Mixup contributes to an additional 0.4\% F1 on the small-size HuSHeM dataset.

\vspace{-4mm}
\section{Conclusions}
\vspace{-3mm}
We proposed a new sperm morphology classification framework, \mname, which employs morphological information of sperm head masks. 
\mname generates boundary-refined masks using image priors and incorporates an effective feature fusion method that leverages features from the raw images and their masks. The \textit{Soft Mixup} component was introduced to handle noisy labels and regularize training on small
datasets.
SOTA performance was obtained on two sperm morphology datasets, without additional pre-training or costly ensembling of multiple CNNs. 
Ablation and component studies demonstrated the effectiveness of \mname.

\vspace{-0.1cm}
\section{COMPLIANCE WITH ETHICAL STANDARDS}
\vspace{-0.1cm}
This research study involved a retrospective analysis of human subject data obtained from two publicly accessible open-access datasets. Ethical approval was not required as confirmed by the licenses attached within the open
access of the two datasets \cite{goldScian, shakerAPDL} used in the experiments of our method.

\bibliographystyle{plain}
\bibliography{refs}{}

\begin{thebibliography}{10}

\bibitem{goldScian}
V.~Chang, A.~Garcia, N.~Hitschfeld, and S.~H{\"a}rtel.
\newblock Gold-standard for computer-assisted morphological sperm analysis.
\newblock {\em CBM}, 83:143--150, 2017.

\bibitem{violetaCESVM}
V.~Chang, L.~Heutte, C.~Petitjean, S.~H{\"a}rtel, and N.~Hitschfeld.
\newblock Automatic classification of human sperm head morphology.
\newblock {\em CBM}, 84:205--216, 2017.

\bibitem{vchang2014goldSeg}
V.~Chang, J.M. Saavedra, V.~Casta{\~n}eda, L.~Sarabia, N.~Hitschfeld, and S.~H{\"a}rtel.
\newblock Gold-standard and improved framework for sperm head segmentation.
\newblock {\em Computer Methods and Programs in Biomedicine}, 117(2):225--237, 2014.

\bibitem{morphSubjectivity}
M.~Freund.
\newblock Standards for the rating of human sperm morphology. a cooperative study.
\newblock {\em International Journal of Fertility}, 11(1):97--180, 1966.

\bibitem{resnet}
K.~He, X.~Zhang, S.~Ren, and J.~Sun.
\newblock Deep residual learning for image recognition.
\newblock In {\em CVPR}, pages 770--778, 2016.

\bibitem{densenet}
G.~Huang, Z.~Liu, L.~Van Der~Maaten, and K.Q. Weinberger.
\newblock Densely connected convolutional networks.
\newblock In {\em CVPR}, pages 4700--4708, 2017.

\bibitem{fusionSCIDS}
H.O. Ilhan and G.~Serbes.
\newblock Sperm morphology analysis by using the fusion of two-stage fine-tuned deep networks.
\newblock {\em Biomedical Signal Processing and Control}, 71:103246, 2022.

\bibitem{ilhan2019automaticEllep}
H.O. Ilhan, G.~Serbes, and N.~Aydin.
\newblock Automatic directional masking technique for better sperm morphology segmentation and classification analysis.
\newblock {\em Electronics Letters}, 55(5):256--258, 2019.

\bibitem{iqbalMCHSH}
I.~Iqbal, G.~Mustafa, and J.~Ma.
\newblock Deep learning-based morphological classification of human sperm heads.
\newblock {\em Diagnostics}, 10(5):325, 2020.

\bibitem{liuTL}
R.~Liu, M.~Wang, M.~Wang, J.~Yin, Y.~Yuan, and J.~Liu.
\newblock Automatic microscopy analysis with transfer learning for classification of human sperm.
\newblock {\em Applied Sciences}, 11(12):5369, 2021.

\bibitem{shufflenet}
N.~Ma, X.~Zhang, H.~Zheng, and J.~Sun.
\newblock {ShuffleNet V2}: Practical guidelines for efficient {CNN} architecture design.
\newblock In {\em ECCV}, pages 116--131, 2018.

\bibitem{menkveld2011measurement}
R.~Menkveld, Cas~A. Holleboom, and J.P Rhemrev.
\newblock Measurement and significance of sperm morphology.
\newblock {\em Asian Journal of Andrology}, 13(1):59, 2011.

\bibitem{whoHandbook}
World~Health Organization et~al.
\newblock {WHO} laboratory manual for the examination and processing of human semen.
\newblock 2010.

\bibitem{world2023infertility}
World~Health Organization et~al.
\newblock Infertility prevalence estimates: 1990--2021.
\newblock 2023.

\bibitem{park1997segmentation}
K.S. Park, W.J. Yi, and J.S. Paick.
\newblock Segmentation of sperms using the strategic {Hough} transform.
\newblock {\em Annals of Biomedical Engineering}, 25:294--302, 1997.

\bibitem{riordonFTVGG}
J.~Riordon, C.~McCallum, and D.~Sinton.
\newblock Deep learning for the classification of human sperm.
\newblock {\em CBM}, 111:103342, 2019.

\bibitem{shakerAPDL}
F.~Shaker, S.A. Monadjemi, J.~Alirezaie, and A.R. Naghsh-Nilchi.
\newblock A dictionary learning approach for human sperm heads classification.
\newblock {\em CBM}, 91:181--190, 2017.

\bibitem{spencer2022ensembled}
L.~Spencer, J.~Fernando, F.~Akbaridoust, K.~Ackermann, and R.~Nosrati.
\newblock Ensembled deep learning for the classification of human sperm head morphology.
\newblock {\em Advanced Intelligent Systems}, 4(10):2200111, 2022.

\bibitem{yuzkat2021multi}
M.~Y{\"u}zkat, H.O. Ilhan, and N.~Aydin.
\newblock Multi-model {CNN} fusion for sperm morphology analysis.
\newblock {\em CBM}, 137:104790, 2021.

\bibitem{zhang2018mixup}
H.~Zhang, M.~Cisse, Y.N. Dauphin, and D.~Lopez-Paz.
\newblock mixup: Beyond empirical risk minimization.
\newblock In {\em ICLR}, 2018.

\bibitem{charleyCasa}
Y.~Zhang, J.~Zhang, X.~Zha, Y.~Zhou, Y.~Cao, and D.~Chen.
\newblock Improving human sperm head morphology classification with unsupervised anatomical feature distillation.
\newblock In {\em ISBI}, pages 1--5. IEEE, 2022.

\end{thebibliography}

\end{document}